\documentclass[letterpaper,twocolumn,10pt]{article}
\usepackage{usenix2019_v3}

\usepackage{graphicx}
\graphicspath{ {images/} }
\usepackage{amsmath}
\usepackage{amsfonts}
\usepackage{amssymb}
\usepackage{color}
\usepackage{comment}
\usepackage{subcaption}
\usepackage{cite}
\usepackage{tikz}
\usepackage{multirow}
\usepackage{float}
\usepackage{breqn}
\usepackage{booktabs}
\newcommand{\ra}[1]{\renewcommand{\arraystretch}{#1}}

\begin{document}

\date{}

\title{\Large \bf DomainGAN: Generating Adversarial Examples to Attack Domain Generation Algorithm Classifiers}

\author{
{\rm Isaac Corley}\\
Booz Allen Hamilton
\and
{\rm Jonathan Lwowski}\\
Booz Allen Hamilton
\and
{\rm Justin Hoffman}\\
Booz Allen Hamilton
} 


\maketitle

\begin{abstract}
Domain Generation Algorithms (DGAs) are frequently used to generate numerous domains for use by botnets. These domains are often utilized as rendezvous points for servers that malware has command and control over. There are many algorithms that are used to generate domains, however many of these algorithms are simplistic and easily detected by traditional machine learning techniques. In this paper, three variants of Generative Adversarial Networks (GANs) are optimized to generate domains which have similar characteristics of benign domains, resulting in domains which greatly evade several state-of-the-art deep learning based DGA classifiers. We additionally provide a detailed analysis into offensive usability for each variant with respect to repeated and existing domain collisions. Finally, we fine-tune the state-of-the-art DGA classifiers by adding GAN generated samples to their original training datasets and analyze the changes in performance. Our results conclude that GAN based DGAs are superior in evading DGA classifiers in comparison to traditional DGAs, and of the variants, the Wasserstein GAN with Gradient Penalty (WGANGP) is the highest performing DGA for uses both offensively and defensively.
\end{abstract}

\section{Introduction}
\label{intro}
Numerous types of malware utilize Domain Generation Algorithms (DGA) to produce a large amount of pseudo-domains. The malware will attempt to beacon to many or all of these domains attempting to find a usable Command and Control (C2) server. These C2 servers provide the malware with further updates such as gathered intelligence \cite{Woodbridge2016PredictingDG} or are used as a means of exfiltration of sensitive information collected from compromised machines. For the malware to be successful, it only requires that a few domains be registered. Additionally, to cause the malware to completely fail, all domains generated and used by the malware must be blacklisted. This makes the task of combating DGAs difficult because DGA detectors need to maintain a near perfect detection accuracy.

\begin{figure*}[ht]
  \includegraphics[width=\linewidth]{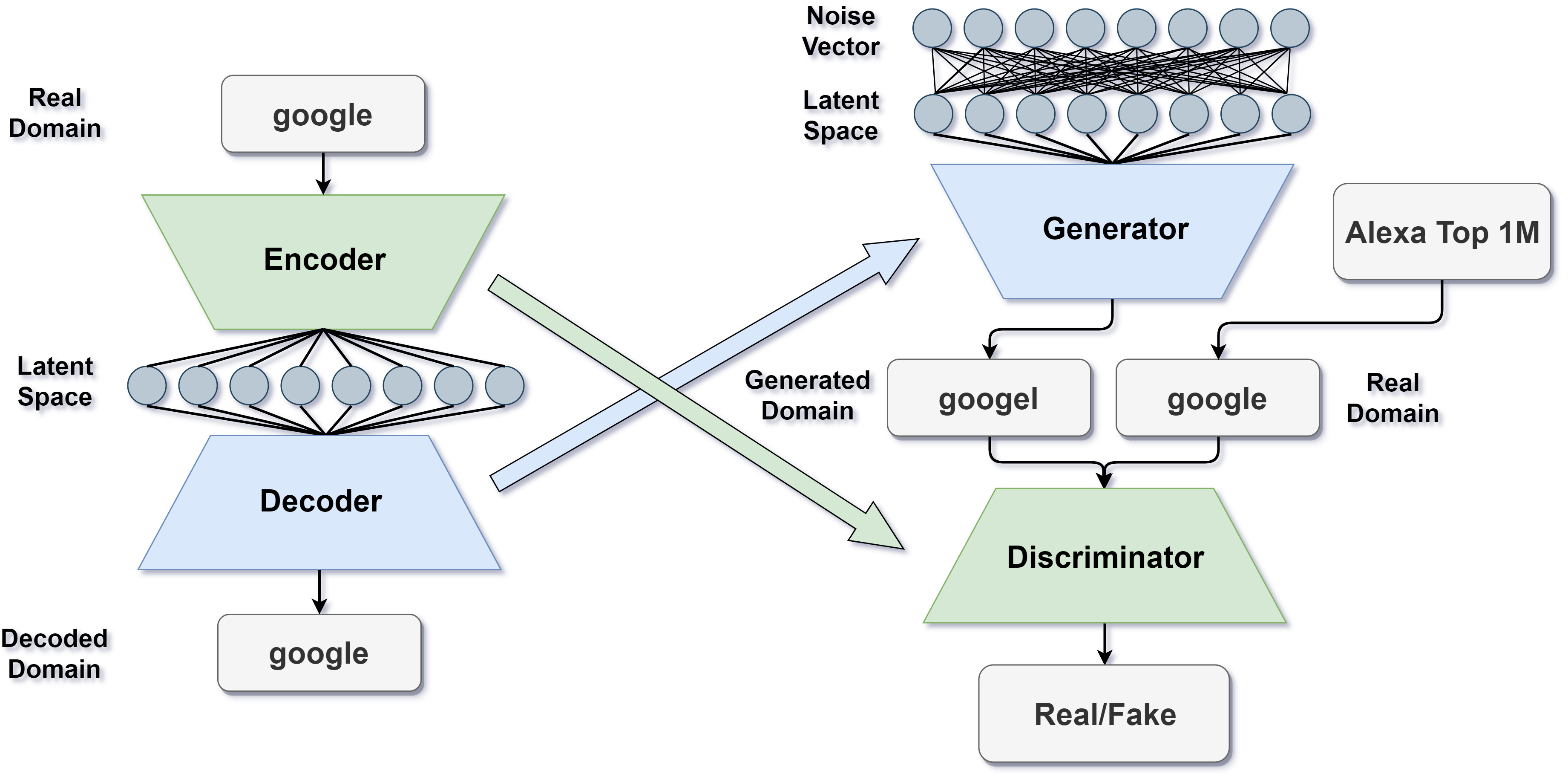}
  \caption{Autoencoder and GAN Architectures}
  \label{fig:architecture}
\end{figure*}

\subsection{Related Work}
Recently, Deep Neural Network (DNN) based DGA classifiers have been developed \cite{Woodbridge2016PredictingDG, yu2018character, Spooren} to attempt to achieve greater performance when detecting DGA created domains, however many of these detection algorithms have only been tested against detecting traditional DGA domains. For example, Woodbridge et. al. developed a DGA classifier using a Long Short Term Memory (LSTM) networks \cite{Woodbridge2016PredictingDG}, \cite{hochreiter1997long}. Their model achieved over 90\% accuracy with a very low false positive rate, however their model was only trained and tested on the Alexa Top 1 Million dataset \cite{alexa}, and the Bambenek DGA domain feeds \cite{Bambenek}. The Bambenek feeds mostly contain domains produced using traditional DGA algorithms. More importantly, the Bambenek feeds are unlikely to contain adversarial DGA domains designed to evade DGA classifiers. Yu et. al. \cite{yu2018character} performed a comparison of state-of-the-art deep learning DGA classifiers which included various Convolutional Neural Network (CNN) \cite{lecun1998gradient}, and LSTM based models. These models were trained on the Alexa Top 1 Million dataset benign domains as well as the Bambenek DGA feeds. Their models resulted in testing accuracies varying from 78\% to 98\%. However, since these models were only trained on the Bambenek feeds, they suffer from the same issues as Woodbridge et. al, e.g. being vulnerable to adversarial examples. 

With the improvement in DGA classifiers, adversarial DGAs have become prevalent \cite{sidi2019maskdga, Peck2019CharBotAS} and developed specifically with the focus of evading maching learning based DGA classifiers. For example, Sidi et. al. \cite{sidi2019maskdga} uses a substitute model to algorithmically perturbate generated domains, making them more likely to evade DGA classifiers. They show that their adversarial DGA degrades the accuracy of various DGA classifier from 97\% to 49\%. Another adversarial DGA developed by Peck et. al. \cite{Peck2019CharBotAS} uses an algorithmic method that introduces small typographical errors in domains sampled from a dictionary of benign domains.

With the emergence of neural networks, machine learning based DGAs have been developed \cite{Spooren, anderson2016deepdga} to specifically evade DGA classifier detection. The DGA developed by Spooren et. al \cite{Spooren}, uses feature engineering along with an iterative DGA development process to produce DGAs that can fool DGA classifiers. Anderson et. al. \cite{anderson2016deepdga} developed a generative DGA, DeepDGA, which trains a Generative Adversarial Network (GAN) to model the distribution of the Alexa Top 1 Million dataset and generate samples which are benign-like to evade DGA classifiers. They tested their DGA samples against a Random Forest DGA classifier \cite{breiman2001random}, and showed that their model had a 48\% detection rate versus the original 96\% detection rate on samples generated by traditional algorithmic DGAs. However, one notable drawback to their model is that it tends to produce very short domains \cite{sidi2019maskdga}. Short domains can be costly for botnet use due to being expensive, having a greater likelihood of already being an existing domain, as well as likely already being previously generated by the DGA in use. While it is possible to use other uncommon top-level domains (TLD) as a solution to producing short domains, this is likely to alarm any defensive system and be quickly flagged.

\subsection{Contribution}
Initial experiments in DeepDGA \cite{anderson2016deepdga} left many unanswered questions and additional analysis regarding the effectiveness of GANs as DGAs. Our contributions consist of a greater exploration into the feasibility of generative deep learning based DGAs in practice. In doing so, we analyze the effects of various GAN variants, as opposed to the single variant used in DeepDGA, to improve domain generation by creating domains which are more difficult for machine learning algorithms to distinguish from benign domains. Furthermore we analyze what it means for a generated domain to be usable for offensive use cases by analyzing features such as domain lengths, n-gram distribution comparisons to real domain datasets, the repetitiveness of a generative models, and the likelihood of generating domains which are already registered. To assess evasion performance, generated domains are compared using multiple state-of-the-art deep neural network DGA classifiers to determine which generative models are most likely to fool DGA classifiers running in production environments. As verified by our results and analysis, the Wasserstein GAN with Gradient Penalty (WGANGP) variant, results in the most usable DGA offensively.

The rest of the paper is organized as follows. The dataset to train the DomainGAN is analyzed in Section \ref{dataset}. Our proposed GAN based DGA will be discussed in Section \ref{GAN}, followed by an analysis of the results in Section \ref{results}, and a discussion of offensive and defensive cases as well as possible future work. Finally, the conclusions and future works are discussed in Section \ref{conclusion}.

\section{Dataset}
\label{dataset}
\begin{figure*}[ht]
  \includegraphics[width=\linewidth]{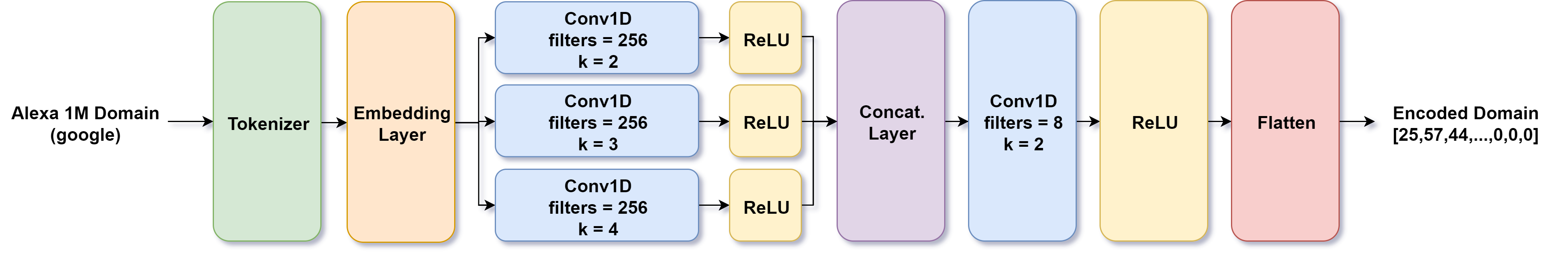}
  \caption{Encoder Architecture for the Autoencoder}
  \label{fig:encoder}
\end{figure*}

\begin{figure*}[!h]
  \includegraphics[width=\linewidth]{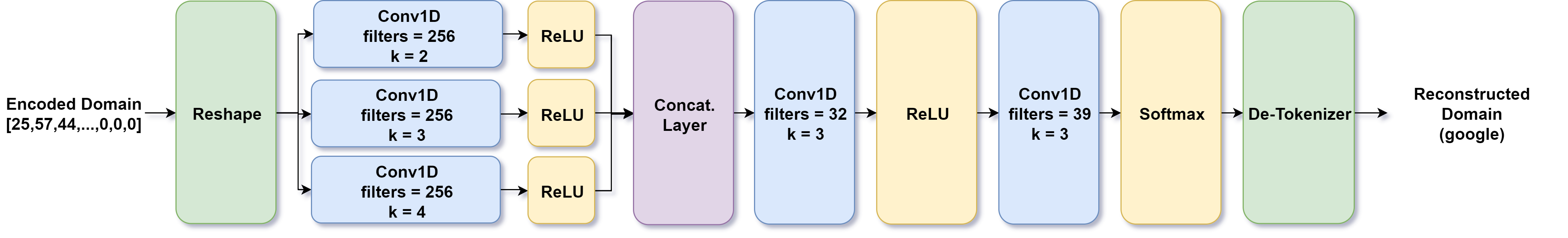}
  \caption{Decoder Architecture for the Autoencoder}
  \label{fig:decoder}
\end{figure*}

The Alexa Top 1 Million dataset \cite{alexa} was used throughout our experiments for generating realistic domain samples. This dataset is composed of the URLs of the top 1 million web sites. The domains are ranked using the Alexa traffic ranking which is determined using a combination of the browsing behavior of users on the website, the number of unique visitors, and the number of pageviews. In more detail, unique visitors are the number of unique users who visit a website on a given day, and pageviews are the total number of user URL requests for the website. However, multiple requests for the same website on the same day are counted as a single pageview. The website with the highest combination of unique visitors and pageviews is ranked the highest \cite{alexa_support}. This ranking provides support to the hypothesis that the Alexa domains are benign domains which are not generated by DGAs. Prior to any experiments, top level domains, e.g. .com, .net, .org, are removed from all domains. To further understand the dataset, a few examples of domains can be viewed in Table \ref{tbl:alexa}.

\begin{table}[!h]
\centering\ra{1.2}
\caption{Alexa Top 1 Million Dataset Examples}
\label{tbl:alexa}
\begin{tabular}{@{}cc@{}}
\toprule
\textbf{Ranking} & \textbf{Domain} \\ \midrule
1 & google.com \\
2 & youtube.com \\
3 & baidu.com \\
... & ... \\
900,000 & aileencooks.com \\
900,001 & alrei.org \\
900,002 & amco.co.in \\ \bottomrule
\end{tabular}
\end{table}

\section{Domain Generation Model}
\label{GAN}
Our proposed GAN model consists of four main components; an encoder, decoder, generator, and discriminator. As seen in Figure \ref{fig:architecture}, the autoencoder is initially trained to take an input domain from the Alexa Top 1 Million dataset, encode that domain into a small finite embedded set of neurons using the encoder network and then decode the compressed representation back into the original domain using the decoder network. After this training process, the autoencoder networks are then rearranged into the GAN framework where the decoder network is repurposed as the generator network and the encoder network is utilized as the discriminator network. The generator is then trained to produce domains which are as similar as possible to the Alexa Top 1 Million domains. The discriminator model then detects if a given domain is produced by either the generator network or sampled from the Alexa Top 1 Million dataset. The generator and discriminator networks will then iteratively learn how to fool and detect the other, respectively. This process is repeated until the generator is able to produce realistic benign-like domains.

\subsection{Autoencoder Model}
Similarly to the experiments of \cite{anderson2016deepdga}, we initialize the generator network's weights by pretraining an autoencoder to learn a compressed representation of important domain specific features in the embedded space. To do this, the autoencoder consists of an encoder, seen in Figure \ref{fig:encoder}, and a decoder, seen in Figure \ref{fig:decoder} both of which are individually inspired by the sentence classification network from \cite{kim2014convolutional}. We note that when not utilizing pretraining, GAN training becomes highly unstable and consistently diverges to unusable samples.

The encoder begins by taking a domain from the Alexa Top 1 Million dataset as input. This domain is then tokenized and fed into an embedding layer with 39 input dimensions representing the set of possible tokens, embedding dimension of 39, and an input sequence length of 60 maximum tokens. The output of the embedding layer is then fed into three parallel 1-dimensional convolutional layers. All three layers have 256 filters and Rectified Linear Unit (ReLU) activations \cite{nair2010rectified}. The three layers have a kernel size of 2, 3, and 4, respectively, which theoretically extracts various n-gram features of the domain names. The 3 parallel convolution layer outputs are then concatenated together and fed into another convolution layer with 8 filters, a kernel size of 2, and a ReLU activation. Finally, the output of the last convolution layer is flattened into a single vector to form the compressed encoder output. This architecture is visualized in Figure \ref{fig:encoder}.

The decoder begins by taking the output of the encoder as its input. The input is then reshaped into a 2-dimensional matrix and fed into 3 parallel convolution layers, similarly to the encoder architecture. The layers' outputs are concatenated together and are fed into another convolution layer. This convolution layer has 32 filters and a kernel size of 3, followed by a ReLU activation. The decoder's final convolution layer is then trained to reproduce the original domain which was fed to the encoder. This layer has 39 filters, a kernel size of 3, and softmax activation. The softmax activation output represents the probability distribution across tokens. This architecture is visualized in Figure \ref{fig:decoder}.

\subsection{Generator Model}
\begin{figure}[!h]
  \includegraphics[width=\linewidth]{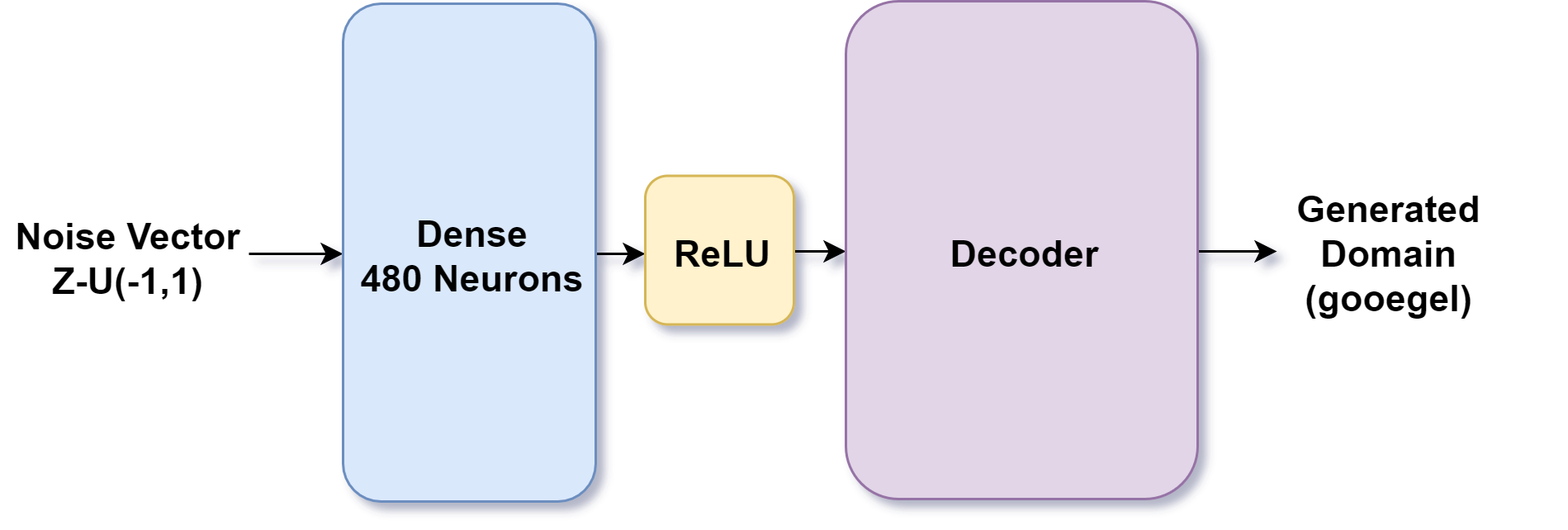}
  \caption{Generator Architecture}
  \label{fig:generator}
\end{figure}

Once the decoder has been trained to learn to decode the low-dimensional representation of benign domains, it is repurposed for use as the generator in the GAN framework. The generator, seen in Figure \ref{fig:generator}, takes a latent vector $z$, sampled from a random uniform distribution on the interval [-1, 1] as its input, or more formally $z$$\sim$$U(a, b)$ where $a=-1$ and $b=1$. This vector is fed into a fully-connected layer with 480 neurons and a ReLU activation. The output of this layer is then fed into the pretrained decoder. The pretrained decoder's weights are frozen, and the output of the decoder is the generated domain. Intuitively, the fully-connected layer learns a mapping from a uniform distribution to the low-dimensional distribution of the embedded space learned by the encoder to produce realistic benign domains. The generator architecture is displayed in Figure \ref{fig:generator}.

\subsection{Discriminator Model}
\begin{figure}[!h]
  \includegraphics[width=\linewidth]{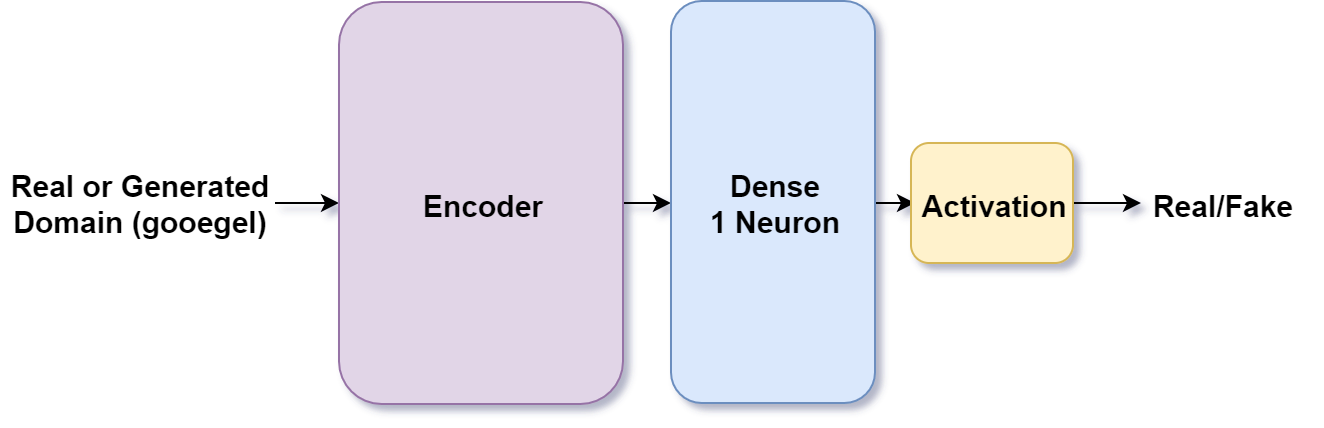}
  \caption{Discriminator Architecture}
  \label{fig:discriminator}
\end{figure}

Similar to the generator, the discriminator is developed using the pretrained decoder weights as its initialization. The discriminator, seen in Figure \ref{fig:discriminator}, takes a domain that is real or generated as the input. The domain is then fed into the pretrained encoder from the autoencoder. The encoder's weights are frozen as well. The output of the encoder is then fed into a single neuron output layer with linear activation. The output of this layer is the probability that the input domain was sampled from the Alexa Top 1 Million or generated. The discriminator architecture is displayed in Figure \ref{fig:discriminator}.

\section{Results}
\label{results}
\subsection{Autoencoder Training Results}

The autoencoder was trained on the Alexa Top 1 Million dataset discussed in Section \ref{dataset}. The dataset is randomly shuffled and split into train and test sets with a percentage split criterion of 75/25. The autoencoder is trained for 400 epochs with a batch size of 64. We then calculate the mean squared error (MSE) on the test set which resulted in a MSE of $4.159 \times 10 ^{-6}$. By sampling the maximum token probability from the softmax output distributions, we note that the autoencoder is able to nearly perfectly recreate the test set domains.


%

\subsection{GAN Variants}
After the autoencoder has been trained, the model is split into the encoder and decoder networks which are then used as components of the the discriminator and generator networks, respectively. To train the GAN we have the generator network produce batches of ``fake" domains with an equivalent number of real domains sampled from the Alexa Top 1 Million dataset. The discriminator then attempts to determine if the domains are fake or real. Based on how well the discriminator is able to classify the domains, the weights of the generator and the discriminator are both updated using a loss function and back propagation. It is known that GANs suffer greatly from instability during training. As a result, convergence during optimization is generally difficult to achieve \cite{mescheder2018training}. To combat this issue, multiple variants of GANs have been developed to improve upon the originally proposed framework. These variants commonly propose new loss functions which are theoretically able to provide a more meaningful metric which can measure the amount the discriminator determines a given sample is real or generated. Our experiments provide an analysis on the task of generating realistic domains by comparing three GAN variants, Least Squares GAN (LSGAN), Wasserstein GAN with Gradient Penalty (WGANGP), and the original GAN, utilized by DeepDGA \cite{anderson2016deepdga}.

The original GAN loss function solves the binary classification problem of determining of whether an input to the discriminator network is either sampled from the real data or generated by the generator network. The output of the discriminator is composed of a sigmoid activation which the output can be derived either 1 (real) or 0 (generated/fake). The objective function is realized in Equation \ref{eq:GAN_Loss}.

\begin{equation}
\label{eq:GAN_Loss}
\begin{split}
    \min_{G}\max_{D}V_{\text{\tiny GAN}}(D,G) = & \mathbb{E}_{\textbf{x}\sim p_{\textit{data}}(\textbf{x})}[\textit{log}D(\textbf{x})] + \\
    & \mathbb{E}_{\textbf{z}\sim p_{\textit{z}}(\textbf{z})}[\textit{log}(1-D(G(\textbf{z})))] \\
\end{split}
\end{equation}

The LSGAN framework \cite{lsgan} was proposed to solve the vanishing gradient problem inherent in neural network classifiers with sigmoid outputs. The modified discriminator output is meant to provide an unbounded measurement of correctness to more effectively penalize the discriminator’s classifications. This change effectively makes the discriminator network a critic instead of a classifier as it's able to provide a value which is more similar to a continuous score than a classification. The notable changes within the GAN framework are the replacement of the discriminator sigmoid output activation with a linear activation and optimizing the discriminator with a MSE loss function. The objective functions for the LSGAN framework are provided in Equations \ref{eq:LSGAN_Loss1} and \ref{eq:LSGAN_Loss2}.

\begin{equation}
\label{eq:LSGAN_Loss1}
\begin{split}
    \min_{D}V_{\text{\tiny LSGAN}}(D)= & \frac{1}{2}\mathbb{E}_{\textbf{x}\sim p_{\textit{data}}(\textbf{x})}[D(x-b)^2]+ \\
    & \frac{1}{2}\mathbb{E}_{\textbf{z}\sim p_{\textit{z}}(\textbf{z})}[(D(G(\textbf{z}))-a)^2]
\end{split}
\end{equation}

\begin{equation}
\label{eq:LSGAN_Loss2}
\begin{split}
    \min_{G}V_{\text{\tiny LSGAN}}(G) = \frac{1}{2}\mathbb{E}_{\textbf{z}\sim p_{\textit{z}}(\textbf{z})}[(D(G(\textbf{z}))-c)^2]
\end{split}
\end{equation}

The final GAN variant we utilize throughout our experiments is the WGANGP framework. The WGANGP framework, seen in Equation \ref{eq:WGANGP_Loss}, utilizes the Earth Mover's distance, or Wasserstein-1, provided in Equation \ref{eq:WGANGP_Critic}. Due to discriminator network's output metric being representing a continuous value, it is commonly referred to as a critic. The critic provides a continuous metric for comparing real and generated samples which is shown to be a more meaningful representation of comparing the data distributions. In addition to the change in loss function, the WGANGP framework uses a Gradient Penalty which constrains the norm of the gradients of the networks to a maximum of 1, provided in Equation \ref{eq:WGANGP_GP}.

\begin{equation}
\label{eq:WGANGP_Loss}
    \min_{G}\max_{D}V_{\text{\tiny WGANGP}}(D,G) = \text{Critic}(D, G) + \text{GP}(D)\\
\end{equation}

\begin{equation}
\label{eq:WGANGP_Critic}
\begin{split}
    \text{Critic}(D, G) =  & \mathbb{E}_{\textbf{x}\sim p_{\textit{data}}(\textbf{x})}[D(\textbf{x})] + \\
    & \mathbb{E}_{\textbf{z}\sim p_{\textit{z}}(\textbf{z})}[1-D(G(\textbf{z}))] \\
\end{split}
\end{equation}

\begin{equation}
\label{eq:WGANGP_GP}
    \text{GP}(D) = \lambda\mathbb{E}_{\hat{x}\sim p_{\hat{x}}(\hat{x})}[(\lVert\nabla_{\hat{x}}D(\hat{x})\rVert_{2}-1)^2]
\end{equation}

To determine which framework provides the most usable samples, we generate 1 million domains using each trained GAN variant and perform several analyses to assess deployment feasibility for botnets.

\subsection{Domain Length Analysis}
\begin{figure*}[ht]
  \includegraphics[width=\linewidth]{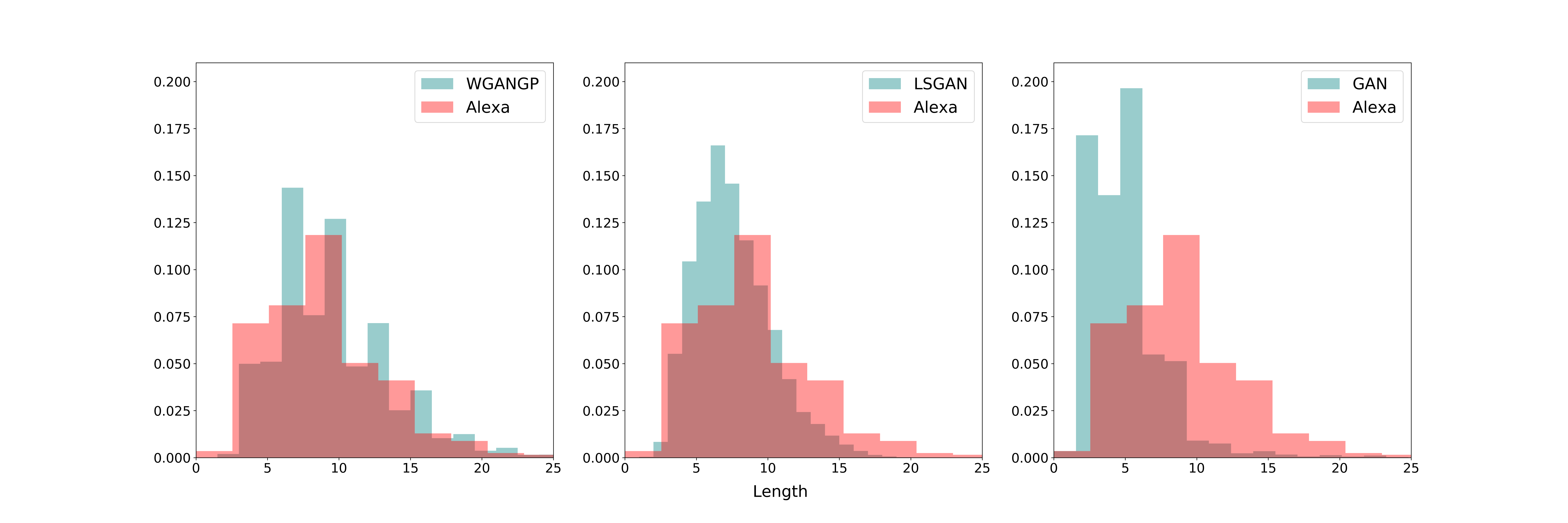}
  \caption{Generated Domain Lengths Distributions}
  \label{fig:lengths_hist}
\end{figure*}

An analysis was performed to compare the domain lengths of the generated domains to the Alexa Top 1 Million domains. Generated samples with lengths similar to benign domains are important for evasion because DGA classifiers will typically learn features such as length of domains to differentiate benign from DGA domains. Additionally, shorter domains increase the likelihood of a domain collision resulting in a more expensive cost to register the domain. An existing domain collision can be defined as the case when a DGA generates a domain which already exists or is owned by another entity. This results in an objective where DGAs should seek to generate samples with similar domain length distributions as that of benign domains. As seen in Figure \ref{fig:lengths_hist}, the original GAN learns to generate notably small domains, even smaller than the Alexa domain length distribution. However, the WGANGP model is more visually similar to the domain length distribution of the Alexa Top 1 Million dataset in comparison to the other GAN variants.

\subsection{Existing Domain Collision Analysis}

To provide further analysis on the effects of domain length on a GAN variant's ability to produce a usable domain, an analysis was performed to calculate the percentage of domains produced by each GAN variant which are already owned. If a given domain already exists then the generated domain is unusable by a botnet unless purchased from the existing owner. To check the performance of each GAN variant with respect to generating unusable existing domains, 1000 domains where generated by each GAN variant and then tested for existence online. Each generated second level domain was concatenated with 3 top level domains, ``.com", ``.org", and ``.net". As seen in Table \ref{tbl:domainexists}, the WGANGP produces significantly less existing domain collisions in comparison to the other GAN variants. The WGANGP produces 12.3\% existing domain collisions, the LSGAN 19.6\%, and the GAN 29.6\%.

\begin{table}[h!]
\centering\ra{1.2}
\caption{Percentage of Generated Domains Resulting in Existing Domain Collisions}
\label{tbl:domainexists}
\begin{tabular}{@{}lc@{}}
\toprule
\textbf{GAN Variant} & \textbf{Existing Domain Collision \%} \\ \midrule
GAN & 29.6 \\
LSGAN & 19.6 \\
WGANGP & \textbf{12.3} \\ \bottomrule
\end{tabular}
\end{table}

\subsection{Repeated Domain Collision}
Another important aspect to consider when comparing a DGA is repeated domain collision. A repeated domain collision can be defined as the likelihood of the DGA to produce the same domain more than once in a batch of generated samples. When generating domains for use offensively, it can become costly to assess whether a domain is in fact usable. To analyze repeated domain collisions, all duplicates were removed from the 1 million generated domains. As seen in Table \ref{tbl:domaincollision}, the original GAN had the highest amount of repeated domain collisions at 53.2\%, while the WGANGP had the lowest amount at 7.4\%. Intuitively, repeated domain collisions can be linked to the domain length distributions of each GAN variant, since shorter domains are likely to have a higher chance of repetition than longer domains. Additionally, the results in Table \ref{tbl:domaincollision} conclude that the WGANGP variant results in minimal repeated domain collisions at 7.4\%, LSGAN at 16.1\%, and GAN at 53.2\%.

\begin{table}[h!]
\centering\ra{1.2}
\caption{Percentage of Generated Domains Resulting in Repeated Domain Collisions}
\label{tbl:domaincollision}
\begin{tabular}{@{}lc@{}}
\toprule
\textbf{GAN Variant} & \textbf{Repeated Domain Collision \%} \\ \midrule
GAN & 53.2 \\
LSGAN & 16.1 \\
WGANGP & \textbf{7.4} \\ \bottomrule
\end{tabular}
\end{table}

\subsection{N-gram Distribution Analysis}
\begin{figure*}[ht]
  \includegraphics[width=\linewidth]{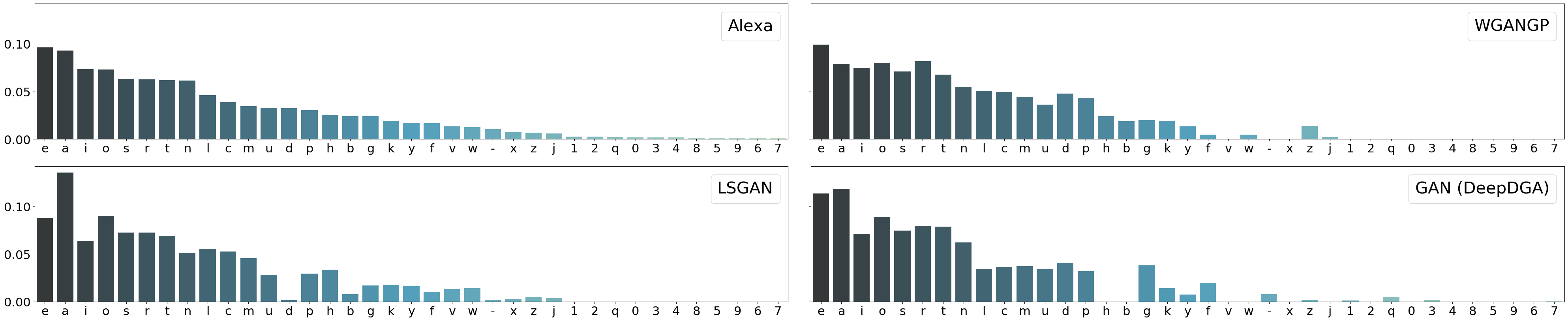}
  \caption{Unigram Character Distributions of Alexa Top 1M and Generated Domains}
  \label{fig:unigrams}
\end{figure*}

\begin{figure*}[ht]
  \includegraphics[width=\linewidth]{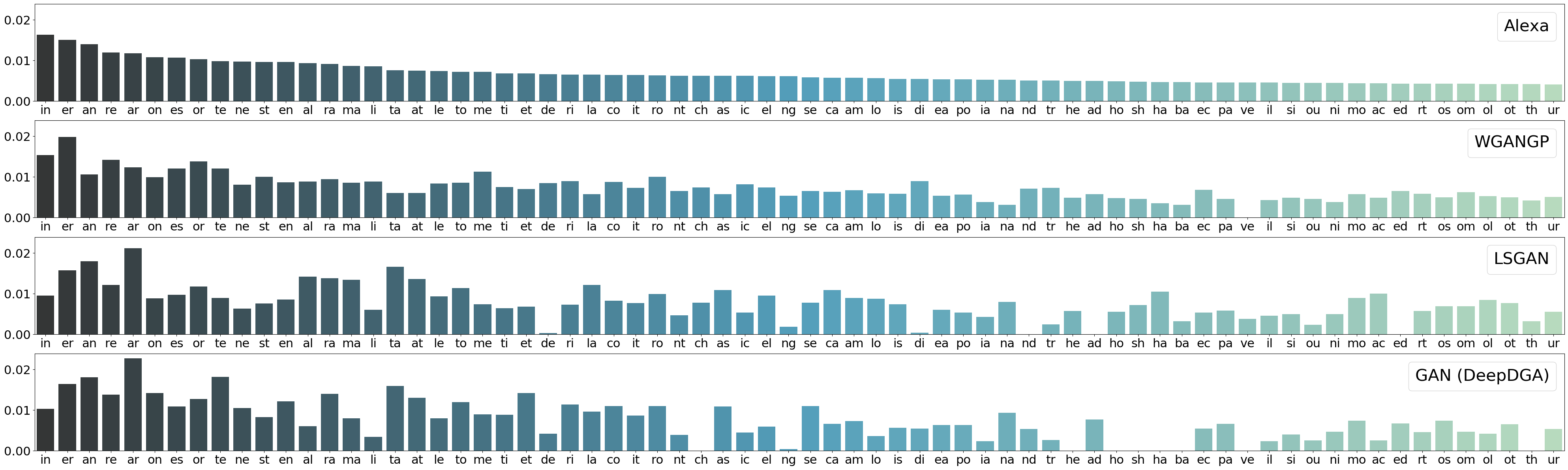}
  \caption{Bigram Character Distributions of Alexa Top 1M and Generated Domains}
  \label{fig:bigrams}
\end{figure*}

To further compare generated and benign samples, the unigram and bigram distributions of each GAN variant's 1 million generated samples are calculated and analyzed. Similarly to domain lengths, DGA classifiers will typically learn n-gram statistics of domains to differentiate between DGA generated and benign domains. Therefore, if a DGA is able to mimic the unigram and bigram character distributions of the Alexa Top 1 Million dataset, it is more likely to evade detection by DGA classifiers trained on the benign samples. As seen in Figure \ref{fig:unigrams} and Figure \ref{fig:bigrams}, we plot the unigram and bigram distributions of the Alexa and generated domains ranked by the Alexa Top 1 Million n-gram distribution in decreasing order. For both n-gram distributions, the WGANGP framework is more notably able to model the Alexa Top 1 Million n-gram distributions in comparison to the LSGAN and GAN variants.

\subsection{DGA Classifier Results}
Furthermore, domains generated by each of the GAN variants were tested against several state-of-the-art DGA classifiers to assess the robustness of the models to the generated adversarial examples. The classifier implementations were sampled from \cite{yu2018character} and are labeled Endgame, Invincea, CMU, MIT, NYU, and Baseline. The original classifiers were then fine-tuned using domains generated from the GAN variants. After fine-tuning these models, the GAN generated domains were again tested to assess the improvement of the fine-tuned DGA classifiers in detecting the adversarial example domains.

\begin{table}[!h]
\centering\ra{1.2}
\caption{Train and Test Accuracies of DGA Classifiers Prior to Fine-Tuning on GAN Generated Samples}
\label{tbl:classifier_results}
\begin{tabular}{@{}lcc@{}}
\toprule
\multicolumn{1}{c}{\textbf{\begin{tabular}[c]{@{}c@{}}Classifier\end{tabular}}} & \multicolumn{1}{c}{\textbf{\begin{tabular}[c]{@{}c@{}}Train Accuracy\end{tabular}}} & \multicolumn{1}{c}{\textbf{\begin{tabular}[c]{@{}c@{}}Test Accuracy\end{tabular}}} \\ \midrule
Endgame & 95.86 & 96.02 \\
Invincea & 98.44 & 98.55 \\
CMU & 95.51 & 95.47 \\
MIT & 98.21 & 98.08 \\
NYU & 98.45 & 98.36 \\
Baseline & 95.49\% & 95.58\% \\ \bottomrule
\end{tabular}
\end{table}

\begin{table}[!h]
\centering\ra{1.2}
\caption{Percentage of GAN Generated Domains to Evade Detection By DGA Classifiers}
\label{tbl:classifier_results_gan}
\begin{tabular}{@{}lccc@{}}
\toprule
\multicolumn{1}{c}{\textbf{Classifier}} & \multicolumn{1}{c}{\textbf{\begin{tabular}[c]{@{}c@{}}GAN\\ Evasion \%\end{tabular}}} & \multicolumn{1}{c}{\textbf{\begin{tabular}[c]{@{}c@{}}LSGAN\\ Evasion \%\end{tabular}}} & \multicolumn{1}{c}{\textbf{\begin{tabular}[c]{@{}c@{}}WGANGP\\ Evasion \%\end{tabular}}} \\ \midrule
Endgame & 98.93 & 95.58 & 96.14 \\
Invincea & 97.43 & 94.94 & 94.93 \\
CMU & 99.23 & 98.84 & 97.63 \\
MIT & 98.90 & 97.65 & 97.78 \\
NYU & 97.74 & 95.58 & 96.14 \\
Baseline & 99.63 & 98.89 & 97.22 \\ \bottomrule
\end{tabular}
\end{table}

\subsection{Spoofing the DGA Classifiers}
The DGA classifiers were initially trained using the Alexa Top 1 million domains as the benign samples and 1 million DGA domains sampled from the Bambenek feeds \cite{Bambenek}. The dataset was then randomly shuffled and split into train and test sets using a 70/30 split criterion. Each classifier was trained for 50 epochs with only the model providing the lowest loss on the test set being saved. We note that our train and test set accuracies were similar to the results by Yu et. al. \cite{yu2018character}. The training and testing accuracies for each model are provided in Table \ref{tbl:classifier_results}.

After training the original DGA classifiers, the 1 million generated domains from each of the GAN variants were classified using each of the classifiers. As seen in Table \ref{tbl:classifier_results_gan}, all of the models fail to classify a majority of the GAN generated domains as DGA. These results conclude that domains generated using the GAN variants would evade the DGA classifiers that are not fine-tuned at a high percentage.

\subsection{Fine-Tuned Classifiers}
Due to the original DGA classifiers resulting in low accuracy at detecting DGA domains sampled from the GAN variants, the models were fine-tuned on the GAN generated domain samples for each of the variants to attempt to create more robust forms of the DGA classifiers. The datasets for fine-tuning included 500,000 domains from the Bambenek feeds, 500,000 domains generated from each of the GAN variants, and 1 million domains from the Alexa Top 1 million dataset. Each of the classifiers were then fine-tuned by retraining each of the models with the weights being initialized with the weights from the original training without the GAN generated samples. Classifiers were fine-tuned for 50 epochs with only the model providing the lowest loss on the test set being saved. As seen in Table \ref{tbl:classifier_results_finetuned}, the classifiers have lower accuracy than the original models, however this is expected because the dataset includes GAN generated domains which are harder to classify due to their similarity to benign domains. However, the classifiers still maintain a relatively high accuracy while being more robust to adversarial examples and more usable defensively than the original classifiers. 

\begin{table}[!h]
\centering\ra{1.2}
\caption{Train and Test Accuracies of DGA Classifiers After Fine-tuning on GAN Generated Samples}
\label{tbl:classifier_results_finetuned}

\resizebox{0.8\linewidth}{!}{%
\begin{tabular}{@{}llcc@{}}
\toprule
\textbf{Classifier} & \textbf{GAN Variant} & \textbf{Train Acc.} & \textbf{Test Acc.} \\ \midrule
Endgame    & GAN                     & 89.77          & 89.64          \\
           & LSGAN                   & 87.16          & 87.34          \\
           & WGANGP                  & \textbf{83.81} & \textbf{83.93} \\
Invincea   & GAN                     & 94.79          & 95.22          \\
           & LSGAN                   & 92.20          & 92.84          \\
           & WGANGP                  & \textbf{91.72} & \textbf{92.58} \\
CMU        & GAN                     & 90.59          & 90.50          \\
           & LSGAN                   & 88.03          & 87.99          \\
           & WGANGP                  & \textbf{84.60} & \textbf{84.47} \\
MIT        & GAN                     & 93.67          & 93.59          \\
           & LSGAN                   & 90.95          & 90.86          \\
           & WGANGP                  & \textbf{88.73} & \textbf{88.71} \\
NYU        & GAN                     & 94.55          & 94.39          \\
           & LSGAN                   & 91.93          & 91.69          \\
           & WGANGP                  & \textbf{90.83} & \textbf{90.63} \\
Baseline   & GAN                     & 81.94          & 81.89          \\
           & LSGAN                   & 79.40          & 79.44          \\
           & WGANGP                  & \textbf{78.14} & \textbf{78.30} \\ \bottomrule
\end{tabular}
}
\end{table}

To test assess the robustness of the fine-tuned classifiers to correctly identify GAN generated domains as DGA domains, 500,000 additional unique domains were generated by each GAN variant and classified using the fine-tuned models. As seen in Table \ref{tbl:classifier_results_finetuned_gan}, the fine-tuned models have greater performance at correctly classifying GAN generated domains making it more difficult for the GAN based DGAs to evade detection. It is also notable that models trained on the original GAN generated samples still resulted in high evasion percentages by domains sampled from the LSGAN and WGANGP variants.

\begin{table}[!ht]
\centering\ra{1.25}
\caption{Percentage of GAN Generated Domains to Evade Detection by Fine-Tuned DGA Classifiers}
\label{tbl:classifier_results_finetuned_gan}
\resizebox{\linewidth}{!}{%
\begin{tabular}{@{}llccc@{}}
\toprule
\textbf{\begin{tabular}[c]{@{}l@{}}Fine-Tuned\\ Classifier\end{tabular}} & \textbf{\begin{tabular}[c]{@{}l@{}}GAN \\ Variant\end{tabular}} & \textbf{\begin{tabular}[c]{@{}l@{}}GAN \\ Evasion \%\end{tabular}} & \textbf{\begin{tabular}[c]{@{}l@{}}LSGAN \\ Evasion \%\end{tabular}} & \textbf{\begin{tabular}[c]{@{}l@{}}WGANGP \\ Evasion \%\end{tabular}} \\ \midrule
Endgame    & GAN                     & 11.01          & 63.92          & \textbf{74.83} \\
           & LSGAN                   & 45.18          & 25.42          & \textbf{72.65} \\
           & WGANGP                  & 56.43          & \textbf{65.76} & 34.45          \\
Invincea   & GAN                     & 3.24           & 63.69          & \textbf{75.60} \\
           & LSGAN                   & 36.01          & 7.82           & \textbf{56.78} \\
           & WGANGP                  & 50.95          & \textbf{52.54} & 11.39          \\
CMU        & GAN                     & 10.84          & 65.40          & \textbf{76.89} \\
           & LSGAN                   & 49.44          & 25.26          & \textbf{74.31} \\
           & WGANGP                  & 66.77          & \textbf{73.24} & 37.90          \\
MIT        & GAN                     & 9.16           & 65.39          & \textbf{79.37} \\
           & LSGAN                   & 41.08          & 16.03          & \textbf{68.97} \\
           & WGANGP                  & 52.36          & \textbf{59.07} & 25.71          \\
NYU        & GAN                     & 6.77           & 65.61          & \textbf{77.93} \\
           & LSGAN                   & 43.44          & 16.30          & \textbf{66.61} \\
           & WGANGP                  & 55.55          & \textbf{63.09} & 23.19          \\
Baseline   & GAN                     & 30.69          & 64.13          & \textbf{76.36} \\
           & LSGAN                   & 64.07          & 43.00          & \textbf{77.63} \\
           & WGANGP                  & \textbf{89.50} & 88.44          & 61.77          \\ \bottomrule
\end{tabular}%
}
\end{table}

\subsection{Summary}
\begin{figure}[!h]
\centering
  \includegraphics[width=\linewidth]{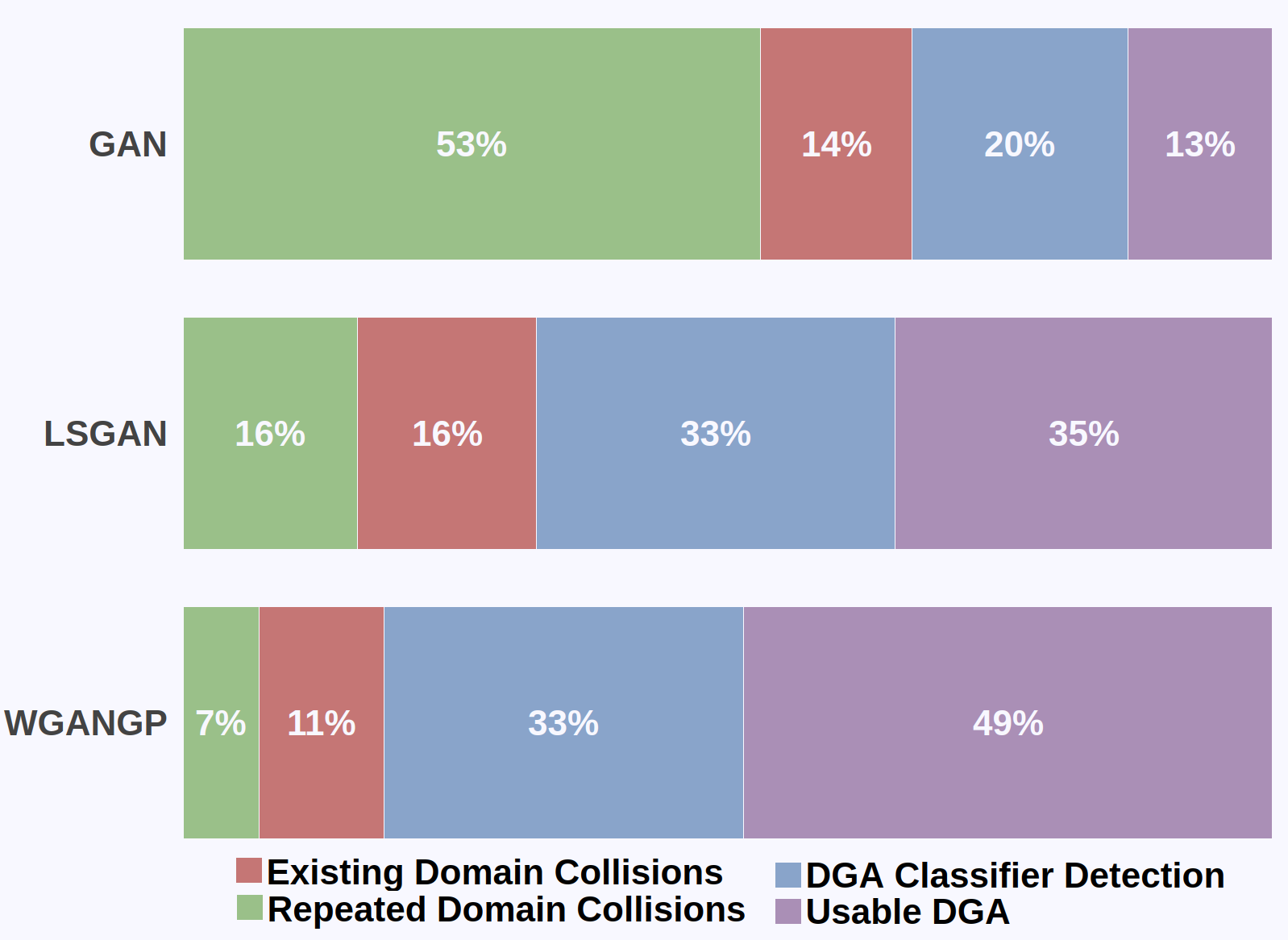}
  \caption{Usability Analysis of Domains Sampled from each GAN Variant}
  \label{fig:domains_stacked}
\end{figure}

To summarize the results in the previous sections, it is necessary to compare the percentage of domains generated from each GAN variant which are actually usable by botnets. The main factors affecting if a given generated domain is usable are ``Repeated Domain Collisions", ``Existing Domain Collisions", and ``DGA Classifier Detections". If a domain encounters any of these issues, it cannot be considered usable for offensive use cases. Using the 1 million generated domains and the DGA classifiers prior to fine-tuning, the probability of a given domain being usable was calculated. Although the WGANGP generated domains have a slightly higher chance of being detected by a DGA classifier, the WGANGP has the highest probability of generating a usable domain. As seen in Figure \ref{fig:domains_stacked}, the WGANGP produces usable domains at a greater rate because it generates significantly less domains which result in a repeated or existing domain collision. It can be concluded that the WGANGP generator is the most usable as a DGA of the compared GAN variants.

\section{Discussion and Future Work}
\label{discussion}
\subsection{Offensive Use Cases}
Using a generative deep learning based DGA could have numerous implementations in practice. As a thought experiment, let us assume we have access to the generative model on both the malware infected machines as well as at the C2 level. Due to the generator model of the GAN being deterministic, a given input noise vector will produce the same output domain. Coordinating the seed with which the noise vector is produced between the compromised machines and the C2 server would allow for predictable rendezvous points for the C2 to utilize. Furthermore, if prior knowledge dictates that a given input noise vector would produce a usable domain, the malware could simply download a list of usable noise vectors with which to produce the adversarial domains.

\subsection{Defensive Use Cases}
The results in Table \ref{tbl:classifier_results} have shown that it is quite simple to evade DGA classifiers trained on traditional DGA domains simply using domains which have similar characteristics to benign domains. This leads to the conclusion that it is actually of great importance to strengthen the decision boundary of any DGA classifiers in production by either fine-tuning on adversarial examples or adding an additional classifier in the pipeline to specifically detect adversarially generated domains.

\subsection{Future Work}
There is much to be explored to improve the performance of generative deep learning based DGAs. Since maintaining similar n-gram characteristics between the DGA domains and benign domains is of importance for evasion, adding an additional objective to the loss function of the GAN, say the Kullback-Leibler Divergence between the softmax output of the generator network and the unigram distribution of the Alexa Top 1M, to assist this would be of interest. Additionally, if possible, adding a constraint to the loss function to penalize short domains, e.g. length < 5, would allow for more usable domains to be generated. Lastly, while we provide a detailed explanation of our convolutional neural network architectures used in the generator and discriminator, we emphasize that this network was simply a baseline to analyze the performance of generative deep learning DGAs as whole and did not perform heavy hyperparameter tuning of the architectures, thus there is possible room for improvement.

\section{Conclusion}
\label{conclusion}
In this paper, three different variants of generative adversarial networks (GANs) are used to improve domain generation by learning the distribution and characteristics of benign domains, making the generated domains more likely to evade detection by state-of-the-art DGA classifiers. Our results conclude that that GAN based DGAs evade detection at a greater rate than traditional DGAs. Additionally, our analysis compared each GAN variant, resulting in the Wasserstein GAN with Gradient Penalty (WGANGP) producing the most usable domains offensively for botnets, due to the low likelihood for repeated and existing domain collisions.

\bibliographystyle{plain}
\bibliography{bibliography}

\end{document}